\documentclass[12pt]{article}
\usepackage{amsmath,amsthm}
\makeatletter
\newcommand\appendix@section[1]{%
  \refstepcounter{section}%
  \orig@section*{Appendix \@Alph\c@section: #1}%
  \addcontentsline{toc}{section}{Appendix \@Alph\c@section: #1}%
}
\let\orig@section\section
\g@addto@macro\appendix{\let\section\appendix@section}
\usepackage{graphicx,psfrag,epsf}
\usepackage{dsfont}
\usepackage{enumerate}
\usepackage{multirow}
\usepackage{natbib}
\usepackage{bm}
\usepackage{bbm}
\usepackage[linesnumbered,ruled,vlined]{algorithm2e}
\newcommand{\blind}{0}
\usepackage{arydshln,leftidx,mathtools}

\usepackage[colorlinks]{hyperref}
\usepackage{xcolor}
\hypersetup{
    colorlinks,
    linkcolor={red!50!black},
    citecolor={blue!80!black},
    urlcolor={blue!80!black}
}
\usepackage{graphicx}
\usepackage{xcolor}
\usepackage[linesnumbered,ruled,vlined]{algorithm2e}
\usepackage{amsfonts}

\SetKwInput{KwInput}{Input}                
\SetKwInput{KwOutput}{Output} 
\usepackage{amsmath}
\DeclareMathOperator*{\argmin}{arg\,min}
\DeclareMathOperator*{\argmax}{arg\,max}
\newcommand{\vect}{\mathtt{vec}}

\newcommand{\E}{{\mathbb{E}}}

\newcommand\smallO{
  \mathchoice
    {{\scriptstyle\mathcal{O}}}
    {{\scriptstyle\mathcal{O}}}
    {{\scriptscriptstyle\mathcal{O}}}
    {\scalebox{.5}{$\scriptscriptstyle\mathcal{O}$}}
  }

\newtheorem{ass}{Condition}

\newtheorem{Thm}{Theorem}
\newtheorem{corollary}{Corollary}

\theoremstyle{definition}
\newtheorem*{example}{Example}
\newtheorem{remark}{Remark}

\newtheorem{definition}{Definition}

\begin{document}

\def\spacingset#1{\renewcommand{\baselinestretch}%
{#1}\small\normalsize} \spacingset{1}


\if0\blind
{
  \title{\bf Regression for matrix-valued data via Kronecker products factorization 
  }
  \author{Yin-Jen Chen\\
    Minh Tang \\
    Department of Statistics, North Carolina State University}
  \maketitle
} \fi

\if1\blind
{
  \bigskip
  \bigskip
  \bigskip
  \begin{center}
    {\LARGE\bf Title}
\end{center}
  \medskip
} \fi

\bigskip
\begin{abstract}
  We study the matrix-variate regression problem
  $Y_i = \sum_{k} \beta_{1k} X_i \beta_{2k}^{\top} + E_i$ for $i=1,2\dots,n$ in the
  high dimensional regime wherein the response $Y_i$ are matrices
  whose dimensions $p_{1}\times p_{2}$ outgrow both the sample size
  $n$ and the dimensions $q_{1}\times q_{2}$ of the predictor
  variables $X_i$ i.e., $q_{1},q_{2} \ll n \ll p_{1},p_{2}$.
  We propose an estimation algorithm, termed KRO-PRO-FAC, for estimating
  the parameters $\{\beta_{1k}\} \subset \Re^{p_1 \times q_1}$ and
  $\{\beta_{2k}\} \subset \Re^{p_2 \times q_2}$ that utilizes the
  Kronecker product factorization and rearrangement operations from
  \cite{Kronecker_Approx}. The KRO-PRO-FAC algorithm is
  computationally efficient as it does not require estimating the
  covariance between the entries of the $\{Y_i\}$.
  We establish perturbation bounds between $\hat{\beta}_{1k} -
  \beta_{1k}$ and $\hat{\beta}_{2k} - \beta_{2k}$ in spectral norm
  for the setting where either the rows of $E_i$ or the columns of
  $E_i$ are independent sub-Gaussian random vectors.  Numerical studies
  on simulated and real data indicate that our procedure is
  competitive, in terms of both estimation error and predictive accuracy, 
  compared to other existing methods.
\end{abstract}

\noindent%

\spacingset{1.45}
\section{Introduction}
\label{sec:intro}
Regression is one of the most important and widely studied
inference tasks in statistics and machine learning. Traditional
applications of regression mainly focus on settings where the
response variables $Y$ are either scalars or, more generally, $Y \in
\Re^{d}$ for some ``small'' $d$. With the
recent advancements in computation and storage technology, it is now
common to encounter scenarios where the responses are (large)
matrices. Examples include data from multivariate
bioassay study \citep{bioassay}, electroencephalography
\citep{tensor_regression}, images denoising
\citep{image_Kronecker_svd,image_Kronecker_Toeplitz,kronecker_approx_image},
and factors models in econometrics
\citep{matrix_factor_Constrained,matrix_factor_econometrics}. 



These type of data naturally leads to the simple and intuitive
notion of matrix-variate regression wherein, given a collection of 
predictor and response tuples $\{(X_{i},Y_{i})\}_{i=1}^{n}$ with
$X_i\in  \Re^{q_{1}\times
  q_{2}}$ and $Y_i \in \Re^{p_{1} \times p_{2}}$, one typically assumes that
the $Y_i$ are related to the $X_i$ through the linear model
\begin{align}\label{eq: intro_vec}
    \vect{(Y_{i})}=\nu \,\vect{(X_{i})}+\vect{(E_{i})},\quad i=1,...,n
\end{align}
where $\nu \in \Re^{p_{1}p_{2}\times q_{1}q_{2}}$ are the unknown
regression coefficients, $E_i \in \Re^{p_1 \times p_2}$ are the
unobserved noise matrices, and $\vect$ denote the vectorization
operator that concatenates the column vectors of the input matrix. We
note that any linear model for $\{(X_i, Y_i)\}$ can be written in the
form of Eq.~\eqref{eq: intro_vec}
%
In the high-dimensional regime wherein the dimensions of response
variables $(p_{1}.p_{2})$ grow much faster than the sample size $n$,
i.e., $p_{i}/n \rightarrow \infty$, the regression coefficient $\nu$
is overparameterized and consistent estimation of $\nu$ is generally
unfeasible unless one impose some structural assumptions on
Eq.~\eqref{eq: intro_vec} so as to reduce the effective number of
parameters in $\nu$. 

Two of the most widely studied and adopted regularity conditions for $\nu$ is that
it is low-rank and/or sparse; see
e.g.,
\cite{low_rank_multi_variat_factor,low_rank_multi_variat_factor_sparserow,low_rank_multi_variat_factor_explain_sparse_rank,low_rank_multi_variat_sparseSVD,low_rank_multi_joint,low_rank_multi_joint_findr,low_rank_multi_variat_SEED,low_rank_multi_variate2020}
and the references therein. In particular
\cite{low_rank_multi_variat_factor_explain_sparse_rank} noted that
low-rank constraints are analogous to imposing sparsity on the
data without explicitly specifying any basis.

Despite the popularity of these sparse and/or low-rank assumptions,
they do not lead to a significant reduction in complexity of $\nu$ when
the feature vectors $\{X_i\}$ are low-dimensional
but the response $\{Y_i\}$ are high-dimensional. More specifically,
suppose $p_{1}=p_{2}=p$, $q_{1}=q_{2}=q$, $p \gg q$ and
$p/n\rightarrow \infty$. If we only assume that $\nu$ is low-rank so
that $\mathrm{rk}(\nu) = d \ll n$ then we still need to estimate
on the order of $O(d(p^2 + q^2))$ parameters for any matrix
factorization of $\nu$ (such as SVD) and is computationally infeasible as it
is equivalent to the estimation of covariance matrices in
high-dimensional univariate linear regression with $p$ predictor
variables and $n$ scalar responses. 
In contrast if we assume sparsity on $\nu$ then, denoting the number of non-zero entries in $\nu$ by $s$, we will
in general need $n = \omega(s)$ to estimate $\nu$ consistently. As
$s/p^2 \ll p^{-1}$, this implies that almost all of the entries in the
responses $\{Y_i\}$ are ignorable. This is a rather strong assumption that
should be justified on a case-by-case basis.  

In this paper we consider a more refined variant of the low-rank
assumption on $\nu$ by assuming that it admits a representation/
approximation in terms of a (sum of) Kronecker products of smaller matrices. More
specifically, we shall assume that $\nu$ is of the form
$$\nu = \sum_{k=1}^{d} \beta_{2k} \otimes \beta_{1k}$$
for some collection of $p_1 \times q_1$ matrices $\{\beta_{11}, \dots,
\beta_{1d}\}$ and $p_2 \times q_2$ matrices $\{\beta_{21}, \dots,
\beta_{2d}\}$. The number of effective parameters in $\nu$ is then
$d(p_1 q_1 + p_2 q_2)$ which is substantially smaller than
$O(d(p_1 p_2 + q_1 q_2))$ for $p_1 \gg q_1, p_2 \gg q_2$. See
\cite{Numerical_Kron_PNAS,Numerical_Kron_multilinear,sum_kronecker_approx}
for futher discussion of Kronecker product factorization and its use
in large-scale matrix approximations. 

Finally the linear model in Eq.~\eqref{eq: intro_vec} with the
Kronecker product structure for $\nu$ is
equivalent to the {\em bi-linear model}
\begin{equation}
  \label{eq:kronecker_intro1}
  Y_i = \sum_{k=1}^{d} \beta_{1k}X_{i}\beta_{2k}^{\top}+E_{i}.
\end{equation}
Under this perspective the $\{\beta_{1k}\}$ (resp. $\{\beta_{2k}\}$) can be interpreted as the
row effects (resp. column effects) of $X_i$ on the response
$Y_i$. The special case of $d = 1$ was considered previously in
\cite{Envelope_Models} wherein the authors studied estimation of
$\beta_{11}$ and $\beta_{21}$ using two-step MLEs; see
Section~\ref{sec:method} for further discussions. In a related vein, 
\cite{matrix_factor_Constrained,matrix_factor_econometrics,matrix_factor_Fan}
considered factor models for $Y_i$ of the form $Y_i = \beta_{1} X_i
\beta_{2}^{\top}$ but, in contrast to the current paper, they assume that the $X_i$ are either unknown or
unobserved. They then propose to estimate $\beta_1$ and $\beta_2$ via
two-step PCA 
Finally, the
bi-linear modeling of $\{Y_i\}$ also arise in the context of image
recognition
\citep{2DPCA_Population_value_decomposition,2DPCA_Neurocomputing,2DPCA_generalized_low_rank,2DPC_face}. In
particular \cite{2DPCA_Population_value_decomposition} proposed the
notion of population value decomposition for summarizing images
population $\{Y_i\}$ by assuming that  $Y_i \approx P V_i D$ where $P$
and $D$ encode ``population frame of reference'' for {\em all}
$\{Y_i\}$ while $V_i$ encode
``subject-level'' features specific to a given $Y_i$. Their $P$ and
$D$ thus serve identical roles to that of $\{\beta_{11},
\beta_{21}\}$ in Eq.~\eqref{eq:kronecker_intro1} (when $d = 1$). 


In this paper we study estimation of $\{\beta_{1k}, \beta_{2k}\}$ for
the model in Eq.~\eqref{eq:kronecker_intro1}. Inspired by the work of
\cite{Kronecker_Approx} on the nearest Kronecker product problem, we
observe that $\nu$ exhibits a low-rank representation after {\em
  reshaping and rearranging} the entries of $\nu$. In other words, while $\nu= \sum_{k=1}^{d} \beta_{2k}\otimes \beta_{1k}$
itself need not be low-rank, its rearranged version still admits a
low-rank representation or approximation. Leveraging this observation
we propose an algorithm, termed KRO-PRO-FAC, for estimating $\nu$ with
computational complexity of $O(p_1 p_2 d q_1 q_2)$ flops; see Section~\ref{sec:method}). We next
studied the theoretical properties of the KRO-PRO-FAC algorithm and
show that it yield, under reasonably mild conditions on the noise $E_i$ and the
dimensions $p_i$ (compared with the sample size $n$), 
consistent estimates of the $\{\beta_{1k}, \beta_{2k}\}$; see
Section~\ref{sec: Theory}. Numerical experiments on simulated and real
data are presented in Section~\ref{sec:numerical}. In particular our
procedure is shown to be competitive, in terms of both estimation error and predictive accuracy, 
to other existing methods.



\section{Methodologies}\label{sec:method}
We now introduce some basic notations used throughout this paper. For $p \in \mathbb{N}$, we denote the set $\{1,...,p\}$ by $[p]$. Let $\mathcal{O}(\cdot)$, $\smallO(\cdot)$ and $\Theta(\cdot)$
represent the standard big-O, little-o and big-theta
relationships. For two arbitrary real sequences $(a_{n})_{n\in
  \mathbb{N}}$ and $(b_{n})_{n\in \mathbb{N}}$, we write $a_{n} \ll
b_{n}$ if $a_{n}/b_{n}$ converges to $0$ as $n \rightarrow \infty$; $a_{n} \asymp
b_{n}$ means $a_{n}/b_{n}$ has a finite and non-zero limit as $n \rightarrow \infty$. For an arbitrary matrix $M=(M_{ij}) \in \Re^{p\times q}$,
the Frobenius norm, spectral norm 
and nuclear norm of $M$ are denoted by 
$\|M\|_{F}$, $\|M\|_{2}$ and $\|M\|_{*}$, and if $M$ is square then
$\mathbf{tr}(M)$ and $|M|$ denote its trace and determinant.
The symbol '$\otimes$' represents the Kronecker product between
matrices while $\bm{I}_{p}$ denote the $p \times p$ identity
matrix. The vectorization of a $p \times q$ matrix $M$ is defined as 
\begin{align*}
\vect(M)=[a_{11}\ \dots\  a_{p1}\ \dots a_{12}\ \dots\ a_{p2}\ \dots\
  a_{p1}\ \dots\ a_{pq}]^T \in \Re^{pq}. 
\end{align*}
and we denote its inverse by $\vect^{-1}(m,p,q)$ where $m$ is a vector
in $\Re^{pq}$. 
\subsection{Dual Kronecker products structure}
We first discuss the special case of Eq.~\eqref{eq:kronecker_intro1}
with $d = 1$, i.e.,  
given a collection of matrix-variate predictors $\{X_{i}\}_{i=1}^{n}
\subset \Re^{q_{1}\times q_{2}}$ and matrix-variate responses
$\{Y_{i}\}_{i=1}^{n} \subset \Re^{p_{1}\times p_{2}}$, we consider the bi-linear model of
the form
\begin{align}
\label{eq: bi-linear basic}
    Y_{i}&=\beta_{1}X_{i}\beta_{2}^{\top}+E_{i}, \quad i \in [n] 
\end{align}
where $\beta_1 \in \Re^{p_1 \times q_1}$ and $\beta_2 \in
\Re^{p_2 \times q_2}$ are the unknown regression coefficients and
$E_i$ are unobserved noise matrices. Under this model each column
(resp. row) of
$Y_i$ is a noisy perturbation of some linear combination of the
columns of $\beta_1$ (resp. rows of $\beta_2^{\top}$).
Recall that Eq.~\eqref{eq: bi-linear basic} can be rewritten in vectors form as
\begin{equation}
\label{eq: bi-linear form each}
    \vect^{\top}{(Y_i)} = \vect^{\top}{(X_i)} (\beta_2^{\top} \otimes \beta_1^{\top})   + \vect^{\top}{(E_i)},\quad i\in [n]
\end{equation}
and thus, by collecting all the $\vect^{\top}{(Y_i)}$ into a matrix
and letting $\nu = \beta_2 \otimes \beta_1$, leads to the linear regression model
\begin{align}
\label{eq: bi-linear form all}
  \mathcal{Y}&=\mathcal{X}\nu^{T}+\mathcal{E}
\end{align}
where $\mathcal{Y}$ and $\mathcal{E}$ are  $n \times p_{1}p_{2}$ 
matrices whose rows are the $\vect^{\top}{(Y_{i})}$ and
$\vect^{\top}{(E_{i})}$ respectively while the design matrix $\mathcal{X}\in \Re^{n \times
  q_{1}q_{2}}$ has rows $\vect^{\top}{(X_{i})}$. Several
variants of formulation in Eq.~\eqref{eq: bi-linear form each} and
Eq.~\eqref{eq: bi-linear form all} have been discussed in the
literature; see e.g., \cite{structured_lasso} for the
case of $p_{1}=p_{2}=1$ and \cite{L2rm_Low_rank} for the case of
$q_{1}=q_{2}=1$ and $\nu$ being low-rank. Here Eq.~\eqref{eq:
  bi-linear form all} assumes that the mean coefficient $\nu$ admits a
Kronecker product representation of $\beta_{2}$ and
$\beta_{1}$. Without the Kronecker product factorization, $\nu$ can be
easily overparameterized with $\mathcal{O}(p_{1}p_{2}q_{1}q_{2})$
elements to be estimated. While $\nu$ is identifiable, is
parametrization in terms of $\beta_{1}$ and $\beta_{2}$ is only
identifiable up to a constant, i.e.,
$\beta_{2}\otimes\beta_{1}=c\beta_{2}\otimes c^{-1}\beta_{1}$ for any
non-zero constant $c$. 

As we allude to in the introduction, the bi-linear model in
Eq.~\eqref{eq: bi-linear basic} had been studied previously in
\cite{Envelope_Models} and we now describe the pertinent details of
this work in the context of the current paper.  
Denote the covariance matrix of $E_{i}$ as  $\Sigma_{\vect{(E)}}$. 
\cite{Envelope_Models} then assume that $\Sigma_{\vect{(E)}}$ can be
decomposed as $\Sigma_{\vect{(E)}}=\Sigma_{2}\otimes\Sigma_{1}$; here
$\Sigma_{1}$ and $\Sigma_{2}$ represent the covariance matrix for the
rows and and columns of $Y_{i}$ respectively,
i.e.,
%
 \begin{align}\label{eq: Kronecker cov}
    \Sigma_{1}=\E\big\{(Y_{1}-\E(Y_{1}))(Y_{1}-\E(Y_{1}))^{T}\big\},\quad \Sigma_{2}=\E\big\{(Y_{1}-\E(Y_{1}))^{T}(Y_{1}-\E(Y_{1}))\big\}
\end{align}
The above structure for 
$\Sigma_{\vect{(E)}}$ is quite natural for longitudinal data
  wherein each subject is measured repeatedly over two different
  domains. For example the rows of $Y_i$ can record measurements over time while the columns of
  $Y_i$ record different covariates. Given Eq.~\eqref{eq: Kronecker
    cov}, the number of parameters in $\Sigma_{\vect{(E)}}$ is then
  reduced drasticically from $\mathcal{O}(p_{1}^{2}p_{2}^{2})$ to
  $\mathcal{O}(p_{1}^{2}+p_{2}^{2})$. This in turn allows the
  dimension  $p_{1}$ and $p_{2}$ to possibly outgrow the sample size $n$, i.e., $n \ll \min\{p_{1},p_{2}\}$. 

\cite{Envelope_Models} then consider MLE estimation of $\beta_1,
\beta_2, \Sigma_1, \Sigma_2$ by further assuming that the $E_{i}$ follows the matrix
normal distribution, i.e., $\vect{(E_{i})} \sim \mathcal{N}(0,
\Sigma_{\vect{(E)}})$; for more on the matrix normal distribution see
\cite{first_matrix_normal,matrix_normal_book} and the references
therein. With
the above Kronecker product structure for the mean and
covariance of $Y_i$, the log-likelihood of $\{Y_{i}\}$ given $\{X_i\}$ is
(ignoring unimportant constants)
\begin{align}\label{eq: log-likelihood}
    2 \ell(\{Y_i\};\theta)&=-np_{1} \ln{|\Sigma_{1}|}
                                                     - np_{2}
                                                       \ln{|\Sigma_{2}|}
                                                       - \sum_{i=1}^{n}\mathbf{tr}\Big\{\Sigma_{2}^{-1}(Y_{i}-\beta_{1}X_{i}\beta_{2}^{\top})^{T}\Sigma_{1}^{-1}(Y_{i}-\beta_{1}X_{i}\beta_{2}^{\top})\Big\}
\end{align}
where $\theta = (\beta_1, \beta_2, \Sigma_1, \Sigma_2)$. Let
$\hat{\theta}$ denote the MLE of $\theta$ from Eq.~\eqref{eq: log-likelihood}. As there are
no closed-form expression $\hat{\theta}$, \cite{Envelope_Models}
proposed a two-stage iterative algorithm for finding $\hat{\theta}$
that is motivated by earlier work of \cite{mle_for_matrix_normal}.
More specifically the algorithm sequentially updates either the row parameters
  $(\beta_{1},\Sigma_{1})$ or the column parameters
  $(\beta_{2},\Sigma_{2})$, with the remaining parameters hold fixed.  
While the dual Kronecker product structure and the resulting MLE procedure
provides a convenient way to model both the mean and covariance of
the rows (and columns) simultaneously, there are nevertheless two
major concerns regarding this approach. Firstly the MLE procedure is
guaranteed to converge only to a {\em stationary} point but not
necessarily a {\em global} optimum. Secondly, the update for
$(\beta_1, \Sigma_1)$ (resp. $(\beta_2, \Sigma_2))$ require inverting
$\Sigma_2$ (resp. $\Sigma_1$) and thus each updates involve possibly
$\mathcal{O}(n(p_1^2 p_2 + p_1 p_2^2))$ flops, which is
computationally prohibitive for moderate and/or large values of $p_1$
and $p_2$.  
In light of the above drawbacks we propose in Section~\ref{sec:kron_low rank} a more computationally
efficient procedure which estimates only $\beta_1$ and $\beta_2$
but not $\Sigma_1, \Sigma_2$ or $\mathrm{Cov}[\vect{(Y_i)}]$. 

\subsection{Kronecker products factorization and low-rank approximation}
\label{sec:kron_low rank}
If we assume a high-dimensional setting where the sample size $n$ is small or
comparable to the dimensions $\min\{p_1, p_2\}$ of the response then
it is generally the case that we can not estimate
$\mathrm{Cov}(Y_i)$ to any reasonable degree of accuracy. One simple
and intuitive remedy to this issue is to ignore the structure in
$\mathrm{Cov}(Y_i)$ and instead focus our effort on estimating $\nu$. 

Our starting point is the observation that although the OLS estimate
$\tilde{\nu} = [(\mathcal{X}^{\top} \mathcal{X})^{-1}\mathcal{X}^{\top}\mathcal{Y}]^{\top}$ is a simple and elegant estimate of $\nu$, it
does not share the same Kronecker product structure as that for $\nu =
(\beta_2 \otimes \beta_1)$. It is thus natural to consider projecting
$\tilde{\nu}$ onto the set formed by Kronecker products of matrices 
with appropriate dimensions. 
In particular let $p_1, q_1, p_2, q_2$ be positive integers and $M$ be a matrix of
dimensions $p_1p_2 \times q_1 q_2$. The nearest Kronecker product
approximation to $M$ with respect to the dimensions $\{p_i, q_i\}$
is defined as
\begin{align}\label{eq:NKP}
    \argmin_{\beta_1 \in \Re^{p_{1}\times q_{1}}, \beta_2 \in \Re^{p_2
  \times q_2}} \|M-\beta_{2}\otimes\beta_{1}\|_{F}^{2}.
\end{align}
\cite{Kronecker_Approx} showed that Eq.~\eqref{eq:NKP} has a
closed-form solution given by the truncated SVD of a {\em rearranged}
version of $M$. 
More specifically first partition $M$ into smaller matrices
$M_{ij}\in \Re^{p_{1}\times q_{1}}$ for $1 \leq i \leq p_2$ and $1
\leq j \leq q_2$, i.e., 
\begin{align}
  M = \begin{bmatrix} 
    M_{11} & M_{12} & \cdots&M_{1q_{2}} \\
    M_{21} & M_{22} & \cdots& M_{2q_2} \\
    \vdots & \vdots & \ddots&\vdots \\
    M_{p_{2}1} & M_{p_{2}2} & \cdots &M_{p_{2}q_{2}}.
    \end{bmatrix}
\end{align}
Next define the rearrangement operation $R(\cdot):\,\Re^{p_{1}p_{2}\times
  q_{1}q_{2}}\rightarrow \Re^{p_{2}q_{2}\times p_{1}q_{1}}$  by
\begin{align}\label{eq:rearrangement}
    R(M)=\begin{bmatrix}
             A_1 \\ A_2 \\ \dots \\ A_{q_2} \end{bmatrix},  \qquad
  A_{j} = \begin{bmatrix} \vect(M_{1j})^{\top} \\ \vect(M_{2j})^{\top}
            \\ \dots \\ \vect(M_{p_2j})^{\top} \end{bmatrix}.
\end{align}
We emphasize that $R(M)$ and $M$ generally have different
dimensions. In particular, if $p_1 \asymp p_2$, $q_1 \asymp q_2$ and
$q_i \ll p_i$ then $M$ is a {\em tall} matrix but the dimensions
of $R(M)$ are comparable. The solution of Eq.~\eqref{eq:NKP} is then equivalent to finding the closest
rank-$1$ representation of $R(M)$, i.e., 
\begin{align}\label{eq: kron_approx_1}
\min_{\beta_1, \beta_2} \|M-\beta_{2}\otimes \beta_{1}\|_{F}=
  \min_{\beta_1, \beta_2} \|R(M)-\vect{(\beta_{2})}\vect{(\beta_{1})}^{\top}\|_{F}
\end{align}
and thus, by the Eckart-Young-Mirsky theorem \citep{low_rank_theorem},
we can take $\beta_{1}=\sigma_1^{1/2}\vect^{-1}(\mathcal{V}_1,p_{1},q_{1})$ and  $\beta_{2}=\sigma_1^{1/2}\vect^{-1}(\mathcal{U}_1,p_{2},q_{2})$
where $\sigma_1, \mathcal{U}_1$ and $\mathcal{V}_1$ 
are the leading singular values and (left and right) singular vectors
of $R(M)$, respectively. See \cite{Kronecker_Approx} for more
details. 
In summary if we assume a Kronecker product structure for the
regression coefficient $\nu$ in Eq.~\eqref{eq: bi-linear form all}
then our estimate for $\nu$, $\beta_{1}$ and  $\beta_{2}$ is given by
\begin{enumerate}[\textbf{\textit{Step}} 1:]
    \item Let $\tilde{\nu}
      \gets
      [(\mathcal{X}^{\top}\mathcal{X})^{-1}\mathcal{X}^{\top}\mathcal{Y}]^{\top}
      \in \Re^{p_1 p_2 \times q_1 q_2}$
      be the OLS estimate of $\nu$ and let $R(\tilde{\nu})$ be its Pitsianis-Van Loan rearrangement; see Eq.~\eqref{eq:rearrangement}.
    \item Compute the SVD 
      $R(\tilde{\nu})=\sum_{k=1}^{r}\hat{\sigma}_{k}\,\hat{\mathcal{U}}_{k}\hat{\mathcal{V}}_{k}^{\top}$
      with $\hat{\sigma}_{1}\geq \dots \geq \hat{\sigma}_{r}$ and $r\leq \min\{p_{1}q_{1},p_{2}q_{2}\}$.
    \item Let $\vect{(\hat{\beta}_{2})}= \hat{\sigma}_1^{1/2}
      \,\hat{\mathcal{U}}_{1}$ and
      $\vect{(\hat{\beta}_{1})}=\hat{\sigma}_1^{1/2} \,\hat{\mathcal{V}}_{1}$
     \item Output the estimate $\hat{\nu}\gets \hat{\beta}_{2}\otimes\hat{\beta}_{1}$ for $\nu$. 
\end{enumerate}
We emphasize that, despite the close connection between Kronecker products approximation
and low-rank approximations described in Eq.~\eqref{eq: kron_approx_1}
, the assumption of a Kronecker
factorization for $\nu$ is quite different from
the assumption that $\nu$ is low-rank. Indeed, the rank of $\nu$ can
be as large as $q_1 q_2$ (assuming $p_1p_2 \geq q_1q_2$) even when
$R(\nu)$ is a rank-$1$ matrix. This difference distinguishes
our work from those which introduce
penalty terms to induce low-rank structure on $\nu$ directly; see 
e.g., \cite{L2rm_Low_rank,time_Low_rank,single_sparse_Low_rank} for
recent examples of this latter approach. 
We now consider a simple simulation study to further illustrate
this distinction. 







\begin{figure}[ht]
\centering
\includegraphics[width=0.87\linewidth]{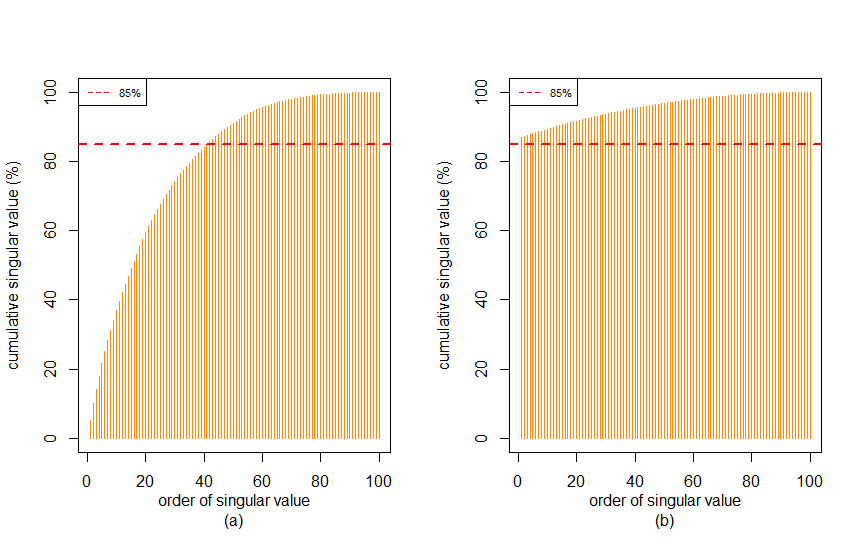}
\caption{\label{fig:cumulative_singular_bar} cumulative singular value
  (averaged over $1000$ replications) for (a) the OLS estimate
  $\tilde{\nu}$ and (b) the rearranged estimate $R(\tilde{\nu})$}
\end{figure}

\begin{example}\label{eg: example1}
We set the dimensions of $Y_{i}$ and $X_{i}$ as
$p_{1}=p_{2}=q_{1}=q_{2}=10$. We then generate $n = 3000$ samples of
the $\{(X_i, Y_i)\}$ pair according to the model $Y_i = \beta_1 X_i
\beta_2^{\top} + E_i$ where the
$\{X_i\}$ are iid random vectors with  $\vect(X_i) \sim
\mathcal{N}(\bm{0}, \mathbf{I})$ and the $\vect(E_i)$ are also iid random
vectors with $\vect(E_i) \sim \mathcal{N}(\bm{0},
\mathbf{I})$. Given the $\{(X_i, Y_i)\}$ we first compute the OLS estimate $\tilde{\nu}$ and its
rearranged version $R(\tilde{\nu})$. Next define, for a matrix $M$ and
an integer $k \geq 1$, the function $f_k(M) = (\sum_{i=1}^{k}
\sigma_i(M))/\|M\|_*$ corresponding to the (normalized) sum of the $k$
largest singular values of $M$; here $\|M\|_*$ denotes the {\em
  nuclear} norm of $M$. 
We then compute, for each $k
\in [100]$, the quantity $f_k(\tilde{\nu})$ and
$f_k(R(\tilde{\nu}))$. Finally we repeat the above steps for $1000$
Monte Carlo replicates. We note that the $\beta_1$ and $\beta_2$ are
fixed constants and do not vary with the Monte Carlo replicates. 
\end{example}
Figure~\ref{fig:cumulative_singular_bar} plots the (normalized)
cumulative sum of the first $k$ singular values of $\tilde{\nu}$ and
$R(\tilde{\nu})$ for $k$ varying in $\{1,2,\dots,100\}$; note that
$f_k(\tilde{\nu}) = f_k(R(\tilde{\nu})) = 1$ when $k = 100$. 
From Figure~\ref{fig:cumulative_singular_bar} we see that the largest
singular value of $R(\tilde{\nu})$ accounts for, on average, roughly
$87\%$ of $\|R(\tilde{\nu})\|_{*}$ and thus a rank-$1$ approximation
of $R(\tilde{\nu})$ is expected to preserve most of the information in
$R(\tilde{\nu})$ while also removing the noise from the small
singular values in $R(\tilde{\nu})$. In contrast the largest singular
value of $\tilde{\nu}$ only explains $5\%$ of
$\|\tilde{\nu}\|_{*}$ and thus computing $\hat{\nu}$ using low-rank
approximations to $\tilde{\nu}$ is possibly problematic. 

\subsection{KRO-PRO-FAC algorithm}
A natural extension of the optimization problem in
Eq.~\eqref{eq:NKP} is to approximate a matrix $\nu$ using a sum of $d$
Kronecker products which, by the above discussions, can be
related to the sum of $d$ rank-$1$ matrices via
\begin{align}\label{eq: kron_approx_d}
    \argmin_{\{(\beta_{1k}, \beta_{2k})\}}
  \|\nu-\sum_{k=1}^{d}\beta_{2k}\otimes \beta_{1k}\|_{F}=
  \argmin_{\{(\beta_{1k}, \beta_{2k})\}}\|R(\nu)-\sum_{k=1}^{d}\vect{(\beta_{2k})}\vect{(\beta_{1k})}^{\top}\|_{F}.\end{align}
A solution of Eq.~\eqref{eq: kron_approx_d} is then once again given
by the truncated SVD of $R(\nu)$. Eq.~\eqref{eq: kron_approx_d}
furthermore suggests a more general version of
the regression problem in Eq.~\eqref{eq: bi-linear basic}, namely that
\begin{align}
    Y_{i}&=\sum_{k=1}^{d}\beta_{1k}X_{i}\beta_{2k}^{\top}+E_{i},\quad
           i\in [n]\label{eq: bi-linear form each d sum}
\end{align}
with $d\ll n \ll \min\{p_{1}q_{1},p_{2}q_{2}\}$. 
Eq.~\eqref{eq: bi-linear form each d sum} can be rewritten as
\begin{align}
\label{eq: bi-linear form all d sum} 
     \mathcal{Y}&=\mathcal{X}\nu^{T}+\mathcal{E},\quad \nu =\sum_{k=1}^{d}\beta_{2k} \otimes \beta_{1k}.
\end{align}
Here we refer to $d$ in Eq.~\eqref{eq: bi-linear form all d sum}, 
as the Kronecker product rank of $\nu$. For ease of exposition (and without
loss of generality) we shall assume that the $\{\beta_{1k}, \beta_{2k}\}$ are
orthogonal, i.e., $\vect{(\beta_{1s})}^{\top} \vect{(\beta_{1t})} = \vect{(\beta_{2s})}^{\top}\vect{(\beta_{2t})}=0$ for all $s \not = t$ and  $\|\beta_{1s}\|_{F}=\|\beta_{2s}\|_{F}$.



Our estimate for $\nu$ and $\{\beta_{1k}, \beta_{2k}\}$ proceeds in an
analogous manner to that described in Section~\ref{sec:kron_low
  rank}. In particular we first compute the OLS estimate $\tilde{\nu}
= [(\mathcal{X}^{\top} \mathcal{X})^{-1} \mathcal{X}^{\top}\mathcal{Y}]^{\top}$, then rearrange $\tilde{\nu}$ to obtain $R(\tilde{\nu})$,
and finally compute the truncated SVD of $R(\tilde{\nu})$ to keep only
the $d$ largest singular values and singular vectors. 
We termed
this procedure as the KRO-PRO-FAC (Kronecker product factorization)
estimate of $\nu$. See
Algorithm~\ref{algo: KRO-PRO-FAC d sum} for a more formal
descriptions. As the Kronecker product rank $d$ of $\nu$ is generally unknown,
we estimate it using the ratio of singular values as described in
\cite{choose_d_AOS} and \cite{choose_d_Eco}, i.e., we estimate $d$ by
\begin{align}\label{eq: choose_d}
    \hat{d}=\argmax_{j\in \{1,\dots,\Bar{d}\}} \hat{\sigma}_{j}/\hat{\sigma}_{j+1}
\end{align}
where $\Bar{d}$ is a pre-specified constant and
$\hat{\sigma}_{k}$'s are the singular values of $R(\tilde{\nu})$ in
a descending order. 

The computational complexity of the KRO-PRO-FAC algorithm is
$\mathcal{O}(p_{1}p_{2}q_{1}q_{2}\min\{p_{1}q_{1},p_{2}q_{2}\})$ with
the main computational bottleneck being the SVD of $R(\tilde{\nu})$.
If $d$ is either known or is estimated to be much
smaller than the dimensions of $R(\tilde{\nu})$ then the cost of the SVD step
reduces to 
$\mathcal{O}(p_{1}p_{2}q_{1}q_{2}d)$ flops by using either Lanczos
bidiagonalization and/or randomized SVD, see e.g.,
\cite{randomized_svd,randomized_svd2,randomized_svd_sketching} and the
references therein. Hence the complexity for the full algorithm itself drops to $\mathcal{O}(np_{1}p_{2}q_{1}q_{2})$. 
In contrast, any algorithm that involves estimating the covariance
matrices for the rows and/or columns will requires at
least $\mathcal{O}(n(p_{1}^{2}p_{2}+p_{1}p_{2}^{2}))$ flops which is an
enormous computational burden for large values of $p_1$ and/or $p_2$. 

\begin{remark}
We note that even if $\nu$ does not have the form as 
specified in Eq.~\eqref{eq: bi-linear form each d sum}
it can nevertheless be well-approximated by a
sum of Kronecker products. Kronecker products provide a
computational efficient building block for approximating
large matrices in numerical linear algebra application. See \cite{Numerical_Kron_PNAS,Numerical_Kron_multilinear,sum_kronecker_approx}
for some general theory and see
\citep{image_Kronecker_svd,image_Kronecker_Toeplitz,cov_Kronecker,cov_Kronecker_robust}
for specific examples in image restoration and covariance
estimation. We emphasize that if $M$
is a $p_1 p_2 \times q_1 q_2$ matrix with $p_1 p_2 \gg q_1 q_2$ then a
rank $d$ SVD of $M$ will require computing left singular vectors
of length $p_1 p_2$ while its Kronecker product factorization 
only require computing factors of dimensions $p_1 \times q_1$ and
$p_2 \times q_2$. 
\end{remark}

\begin{remark} We note that Kronecker products
  factorization also featured prominently in the work of
  \cite{kronecker_approx_image} but their research question is
  susbtantially different from that considered in the current
  paper. In particular our setting is that of linear regression where
  the goal is to estimate the factorization
  $(\beta_{1},\beta_{2})$ of $\nu$ given both the responses 
  $\{Y_{i}\}$ and feature vectors $\{X_{i}\}$, i.e., our estimation of
  $(\beta_1, \beta_2)$ is a {\em supervised learning} problem.
  In contrast \cite{kronecker_approx_image}
  uses Kronecker product approximation to perform dimension reduction
  of the $\{Y_i\}$ without observing any $\{X_i\}$, i.e., they
  are considering an {\em unsupervised} learning problem.  
\end{remark}

\begin{algorithm}\label{algo: KRO-PRO-FAC d sum}
\DontPrintSemicolon
  
  \KwInput{$\mathcal{Y}$, $\mathcal{X}$ and $(p_{2},q_{2})$, $(p_{1},q_{1})$, $\Bar{d}$}
  \KwOutput{$(\Hat{\beta}_{1k}$, $\Hat{\beta}_{2k})$}
     Compute the OLS estimate $\tilde{\nu}\gets[(\mathcal{X}^{\top}\mathcal{X})^{-1}\mathcal{X}^{\top}\mathcal{Y}]^{\top}$.
    \;
    Rearrange $\tilde{\nu}$ to get $R(\tilde{\nu})$ by Eq.~\eqref{eq:rearrangement}.
    \;
    Perform SVD on $R(\tilde{\nu})$, i.e.,
    $R(\tilde{\nu})=\sum_{k=1}^{\Bar{d}}\hat{\sigma}_{k}\,\hat{\mathcal{U}}_{k}\hat{\mathcal{V}}_{k}^{\top}$
    with
    $\hat{\sigma}_{1} \geq \hat{\sigma}_2 \geq \dots \geq \hat{\sigma}_{\bar{d}}$.
    \;
    Estimate $d$ by $ \hat{d}=\argmax_{j\in \{1,\dots,\Bar{d}\}} \hat{\sigma}_{j}/\hat{\sigma}_{j+1}$
      \;
      Set $\vect{(\hat{\beta}_{2k})}= \hat{\sigma}_k^{1/2}
      \,\hat{\mathcal{U}}_{k}$ and
      $\vect{(\hat{\beta}_{1k})}=\hat{\sigma}_k^{1/2} \,\hat{\mathcal{V}}_{k}$. 
     \;
     Output $\hat{\nu} \gets \sum_{k=1}^{\hat{d}} \hat{\beta}_{2k}\otimes\hat{\beta}_{1k}$
\caption{KRO-PRO-FAC algorithm }
\end{algorithm}

\section{Theoretical Results} \label{sec: Theory}
We now study large-sample and/or asymptotic results for the estimates
of $\{\beta_{1k}, \beta_{2k}\}$ obtained by the KRO-PRO-FAC
algorithm. 
Recall that, from our earlier discussions in
Section~\ref{sec:method}, the rearranged OLS estimate $R(\tilde{\nu})$
can be viewed as a sum of rank-$1$ matrices
$R(\nu) = \sum_{k=1}^{d} \vect{(\beta_{2k})}\vect{(\beta_{1k})}^{\top}$ {\em additively perturbed} by the
noise matrix $R(\tilde{\mathcal{E}})$ where $\tilde{\mathcal{E}} =
[(\mathcal{X}^{\top} \mathcal{X})^{-1}\mathcal{X}^{\top}\mathcal{E}]^{\top}$. Therefore, if $\|\tilde{\mathcal{E}}\|$ is sufficiently
small compared to $\|R(\nu)\|$, then we can apply classical results in
matrix perturbation theory such as the
$\sin$-$\Theta$ theorem \citep{wedin_perturbation} to show that 
the leading singular vectors of $R(\tilde{\nu})$ are ``close'' to the
$\vect{(\beta_{1k})}$ and $\vect{(\beta_{2k})}$.

We now make the above description precise. 
Let $R(\nu)$ be a rank $d$ matrix for some fixed constant $d$ not depending on $p_1, p_2$ and $n$. 
Denote the SVD of $R(\nu)$ by $R(\nu) = \mathcal{U} \mathcal{D} \mathcal{V}^{\top}$ where
$\mathcal{D}=diag(\sigma_{k})$ is a $d \times d$ diagonal matrix of singular values, $\mathcal{V}=(\mathcal{V}_{1},\dots,\mathcal{V}_{d})$ is a $p_{1}q_{1} \times d$
orthonormal matrix of right singular vectors and
$\mathcal{U}=(\mathcal{U}_{1},\dots,\mathcal{U}_{d})$ is a $p_{2}q_{2}
\times d$ orthonormal matrix of left singular vectors. 
Next let $\hat{\mathcal{U}}\hat{\mathcal{D}}\hat{\mathcal{V}}^{\top}$
denote the {\em truncated} SVD corresponding to the $d$ largest singular values
and singular vectors of $R(\tilde{\nu})$. 
We first make an assumption on the
relationship between the matrix dimensions $p_{1}$, $p_{2}$, $q_{1}$,
$q_{2}$ and the sample size $n$ as well as the growth rate for the
singular values of $R(\nu)$. 
\begin{ass} \label{ass: asymptotics}
  Let $p_1, p_2$, $q_{1},q_{2}$ and $n$ satisfy
\begin{align*} 
    \frac{q_{1}}{q_{2}} = \Theta(1), \quad \frac{p_{1}}{p_{2}} = \Theta(1),\quad q_{1}q_{2} \ll n, \quad \ln{p_{i}}=o(n),\;i=1,2.
\end{align*}
Furthermore, for sufficiently large $p_{1},p_{2}$, assume that the singular values of $R(\nu)$ satisfy
\begin{align*}
    \sigma_{k} =\mathcal{O}(p_1), \;i=1,2,\dots,d
\end{align*}
\end{ass}
Condition~\ref{ass: asymptotics} 
implies that $R(\nu)$ have bounded condition number. 

We next recall the notion of a sub-Gaussian random vector
\begin{definition}
  Let $Z$ be a mean zero random variable. Then $Z$ is said to be
  sub-Gaussian with variance proxy $\sigma^{2}$ if, for all $t > 0$ we have
\begin{align}\label{eq: sub-Gaussian}
\mathbb{P}(|Z| > t) \leq 2 \exp\bigl(-\tfrac{t^2}{2\sigma^2} \bigr).
\end{align}\end{definition}
In other words, the tail probability of $Z$ behaves similarly to that
of a Gaussian distribution with variance $\sigma^2$.  A mean zero
random vector $\bm{Z} \in \Re^{p}$ is then said to be a sub-Gaussian
random vector with covariance proxy $\Sigma$ if $w^{\top} \bm{Z}$ is
sub-Gaussian with variance proxy $w^{\top} \Sigma w$ for all $w \in
\Re^{p}$.
See Section~2.5
and Section~3.4 of \cite{HDProbability} for further discussion and
characterizations of sub-Gaussian random vectors. 

Now let $\{\xi_{1},\dots,\xi_{n}\}$ be iid mean zero sub-Gaussian
random vectors in
$\Re^{p_{1}p_{2}}$ with covariance proxy $\mathcal{I}$ where
$\mathcal{I}$ is the $p_1 p_2 \times p_1 p_2$ identity matrix. We
shall assume that the noise matrices $E_i$ are of the form 
\begin{align}
\label{eq:noise_Ei}
\vect(E_i)&=\Sigma_{\vect{(E)}}^{1/2}\;\xi_{i},\qquad i \in [n]
\end{align}
for some $p_1 p_2 \times p_1 p_2$ positive definite matrix
$\Sigma_{\vect{(E)}}^{1/2}$ satisfying the following condition. 
\begin{ass}\label{ass: dist and cov}  $\Sigma_{\vect{(E)}}$ is a block
  diagonal matrix , i.e.,
  $\Sigma_{\vect{(E)}}=diag(\bm{\Sigma}_{1},\bm{\Sigma}_{2}, \dots
  ,\bm{\Sigma}_{p_{2}})$ where each diagonal block is of size ${p_{1}
    \times p_{1}}$. Furthermore there exists a positive constant $\mathcal{C}$ independent of $p_{1}$, $p_{2}$ and $n$ such that 
\begin{align}
     \max_{k \in [p_{2}]} \max_{s \in [p_{1}]}\bm{\Sigma}_{k}(s,s) \leq \mathcal{C}.
\end{align}
where $\bm{\Sigma}_{k}(s,t)$ is the $(s,t)$ entry of  $\bm{\Sigma}_{k}$.
\end{ass}

\begin{remark}
We note that the block diagonal structure posited in
Assumption~\ref{ass: dist and cov} is different from and
arguably more flexible than assuming a Kronecker product
structure for $\Sigma_{\vect{(E)}}$. More specifically an arbitrary
$\Sigma_{\vect{(E)}}$ has $O(p_1^2 p_2^2)$ parameters. If
$\Sigma_{\vect{(E)}}$ can be factored into the Kronecker product of a $p_1 \times
p_1$ matrix and a $p_2 \times p_2$ matrix then the number of
parameters is reduced drastically to $O(p_1^2 +
p_2^2)$ parameters.
It was noted in \cite{Kron_cov_complex} that $O(p_1^2 +
p_2^2)$ parameters is potentially too few as it preclude the use
of some common matrix-variate  Gaussian distribution to model
$\Sigma_{\vect{(E)}}$.
In contrast, under
Assumption~\eqref{ass: dist and cov}, $\Sigma_{\vect{(E)}}$ has $O(p_2
p_1^2)$ parameters. If $p_1 \asymp p_2 \asymp p$ then the above three
scenarios correspond to $O(p^4), O(p^2)$ and $O(p^3)$ parameters, respectively. 
Finally we note that Assumption~\ref{ass: dist and cov} is satisfied
whenever the columns of $E_{i}$ are uncorrelated. A similar condition can be formulated for the case when the {\em rows} of $E_i$ are uncorrelated. 
These conditions are milder than assuming that the entries of $E_i$ are
mutually independent as is done in \cite{low_rank_multi_variat_factor_sparserow,low_rank_multi_variat_factor_explain_sparse_rank,low_rank_multi_joint,low_rank_multi_joint_findr,low_rank_multi_variate2020}.
\end{remark}

With the above assumptions in place, 
we now state our theoretical results for bounding the estimation error
between $\hat{\mathcal{U}}$ (resp. $\hat{\mathcal{V}}$) and
$\mathcal{U}$ (resp. $\mathcal{V}$). These errors are
stated in terms of the $\sin$-$\Theta$ distance between linear
subspaces, i.e., given two orthonormal matrices $\mathcal{W}_1$ and
$\mathcal{W}_2$ 
the sin-$\Theta$ distance between the linear subspaces spanned by $\mathcal{W}_1$ and $\mathcal{W}_2$ is defined as 
\begin{align}\label{eq: sin_dist_def}
\|\sin \Theta\:(\mathcal{W}_1,\mathcal{W}_2)\|=\sqrt{1-\sigma_{\min}^{2}(\mathcal{W}_1,\mathcal{W}_2)}.
\end{align}
where $\sigma_{\min}(\mathcal{W}_1, \mathcal{W}_2)$ is the minimum
singular value of $\mathcal{W}_{1}^{\top} \mathcal{W}_2$. 
\begin{Thm}\label{thm: sin_theta} Let $\{(X_i, Y_i)\}$ satisfy the
  linear model in Eq.~\eqref{eq: bi-linear form each d sum} for some
  fixed $d \geq 1$ not depending on $n$ and suppose
  that Condition~\ref{ass: asymptotics} and \ref{ass: dist and cov} holds. 
  Then there exists a constant $C>0$ such that, with probability at least
  $1-n^{-3}$, the following holds {\em simultaneously}, 
\begin{align}
	  |\hat{\sigma}_{k}-\sigma_{k}|&\leq C\,q_{1}q_{2}\max_{k \in p_{2}}\max_{s \in[p_{1}]}\bm{\Sigma}_{k}(s,s)\frac{\sqrt{p_{1}}+\sqrt{p_{2}}+\ln{p_{1}}+\ln{p_{2}}}{\sqrt{n}},\\
   \max\{ \lVert \sin  \Theta\:(\mathcal{U},\hat{\mathcal{U}})\rVert,\, \lVert \sin \Theta\:(\mathcal{V},\hat{\mathcal{V}})\rVert\}&\leq C\,q_{1}q_{2}\max_{k \in p_{2}}\max_{s \in[p_{1}]}\bm{\Sigma}_{k}(s,s)\frac{\sqrt{p_{1}}+\sqrt{p_{2}}+\ln{p_{1}}+\ln{p_{2}}}{\sqrt{n p_{1} p_{2}}}.
\end{align}
\end{Thm}
Theorem~\ref{thm: sin_theta} implies the following upper bound for the
error of $\hat{\nu}=\sum_{k=1}^{d}\hat{\beta}_{2k}\otimes \hat{\beta}_{1k}$
as an estimate for $\nu = \sum_{k=1}^{d} \beta_{2k} \otimes
\beta_{1k}$. In particular the {\em relative} error of $\hat{\nu} - \nu$ converges to $0$ as $p_1, p_2$ and $n$ diverge and thus $\hat{\nu}$ is a consistent estimate for $\nu$. 
\begin{corollary}\label{prop: Kronecker}
Suppose $p_{1}=p_{2}=p$ and consider the setting in Theorem~\ref{thm: sin_theta}. 
Then there exists a constant $C>0$ such that with probability at least $1-n^{-3}$,
\begin{align}
\frac{\|\hat{\nu}- \nu\|_{F}}{\|\nu\|_{F}} \leq \frac{C}{\sqrt{np}}.
\end{align}
\end{corollary}

\section{Numerical experiments}\label{sec:numerical}
We evaluate the numerical performance of the KRO-PRO-FAC algorithm
through a few simulation studies and real data analysis. 
\subsection{Simulation studies} 
\label{subsec:simulation}
For the simulation experiments we set the dimensions of $Y_{i}$ and
$X_{i}$ to be $p_{1}=p_{2}=500$ and $q_{1}=q_{2}=2$ while
the sample size $n$ is chosen in $\{200,400,1000,2000,3000\}$. For
ease of exposition we only consider the special case of 
Eq.~\eqref{eq: bi-linear form each d sum} with $d=1$, and thus $\nu =
\beta_2 \otimes \beta_1$ where $\beta_1$ and $\beta_2$ are $500 \times
2$ matrices. 
We first generate $\vect{(\beta_1)}$ from the standard
multivariate normal distribution on $\Re^{1000}$ and similarly
for $\vect{(\beta_2)}$; note that neither $\beta_1$ nor $\beta_2$
are expected to be sparse and furthermore the estimation of these
$\beta_k$ when $n = 200$ or $n = 400$ falls within the setting of
regression with high-dimensional responses.  We then generate $X_1, X_2, \dots, X_n$
where the $\vect{(X_i)}$ are iid standard multivariate normals in
$\Re^{4}$. 

Given the $\{X_i\}$ we then consider the following $4$ different models for the
random noises $\{E_i\}$. The first three models corresponds to
$\vect{(E_i)} \in \Re^{25000}$ that are multivariate normals while the last
model corresponds to $\vect{(E_i)}$ with entries independently sampled from Student's $t$ distribution with $5$ degrees of freedom. The entries of $E_i$ for Model $4$ have heavier tails compared to that for Models
$1$--$3$. 

\begin{enumerate}[\textbf{\textit{Model}} 1:]
\item \emph{Identity covariance}: $\Sigma_{\vect{(E)}}=\mathbf{I}_{p_{1}p_{2}\times p_{1}p_{2}}$ and $\vect{(E_{i})}$'s are generated independently from the standard multivariate normal distribution. 

\item \emph{Banded covariance}: $\Sigma_{\vect{(E)}}=\mathbf{L}\mathbf{L}^{\top}$ where $\mathbf{L}$ is a lower triangular banded matrix in $\Re^{p_{1}p_{2}\times p_{2}p_{2}}$ with $\mathbf{L}_{ij}=0$ for $i<j$ or $i-j>b$. The bandwidth $b$ is set to $5$ and the diagonal elements are generated from $\mathcal{N}(3,1)$ and the non-zero off-diagonal elements are generated from $\mathcal{N}(0,1)$. $\mathbf{L}$ is fixed over the $100$ replications. 

\item \emph{AR(1)}: $\Sigma_{\vect{(E)}}=(\rho^{|i-j|})_{p_{1}p_{2} \times p_{1}p_{2}}$ with $\rho=0.9$. Here we generate $\vect{(E_{1})},\dots, \vect{(E_{n})}$ based on $\mathtt{Matlab}$ codes from \href{https://www.mathworks.com/help/econ/arima.html;jsessionid=f028e7a0682dd63bcef0ef828c68}{arima}.

\item \emph{Heavy-tailed}: 
$\Sigma_{\vect{(E)}}$ is proportional to $\mathbf{I}_{p_{1}p_{2}\times p_{1}p_{2}}$ and
the entries of $\vect{(E_{i})}$'s are random samples from the Student's t-distribution with $5$ degrees of freedom.
\end{enumerate}

For each choice of the noise model for $E_i$ we then generate
$\{Y_i\}$ according to Eq.~\eqref{eq: bi-linear basic} and then
estimate $\hat{\nu}$ based on the $\{X_i, Y_i\}$ using the KRO-PRO-FAC
algorithm. For illustrative comparisons we considered, in addition to
the default described in Algorithm~\ref{algo: KRO-PRO-FAC d sum}, two other
variants which performs rank regularization of either the responses or
the OLS estimate. More specifically the first variant
uses, instead of the observed $Y_{i}$, its truncated rank$-\alpha$ SVD $Y_i^{(\alpha)}$
for estimating $\nu$. We termed this variant as KRO-PRO-FAC $(\alpha)$
and note that it is motivated by the fact that while
$Y_i$ is, with probability $1$ full rank, 
$\mathbb{E}[Y_i] = \beta_1 X_i \beta_2^{\top}$ is low-rank
for all $i$ and thus a rank-regularized version of the $\{Y_i\}$ might
lead to better estimate of $\nu$. The second variant also performs
rank regularization, but on the OLS estimate $\tilde{\nu}$ as opposed
to the responses $\{Y_i\}$. Letting $\tilde{\nu}^{(\gamma)}$ be the
truncated rank$-\gamma$ SVD of $\tilde{\nu}$ we then perform the
remaining steps of Algorithm~\ref{algo: KRO-PRO-FAC d sum} with
$\tilde{\nu}^{(\gamma)}$ in place of $\tilde{\nu}$. We termed this
variant as rdu-rank-KRO ($\gamma$) and note that it is
motivated by the notion of reduced-rank-regression in
\cite{reduced_rank_regression}. For this simulation we chose $\alpha =
\gamma = 2$.

Finally we also estimate $\nu$ using the 
MLE based procedure described in \cite{Envelope_Models}. Recall
that this MLE based approach posits both a Kronecer product structure for both the
regression coefficient $\nu$ and the covariance matrix of
$\vect{(E_i)}$. We use the implementation from is based on $\mathtt{R}$ codes from \href{https://sites.google.com/a/udel.edu/sding/software}{MatrixEnv}
and denote the resulting estimates as dual-KRO-MLE. 
Table~\ref{tab:method_compare} summarizes some key differences between
the $4$ methods described above. For numerical comparisons 
we evaluate the relative errors 
$\|\hat{\nu}-\nu\|_{F}/\|\nu\|_{F}$ for each methods and {\em
  averaged} these over $100$
independent Monte Carlo replicates. The results are presented in
Table~\ref{tab:M1} through Table~\ref{tab:M4} for the four noise
models described above. 


\begin{table}[!ht]
 \caption{\label{tab:method_compare}Method Comparison}
    \centering
    \begin{tabular}{l|c|c|c}
    \hline
        method&data&$\nu$ estimation& Kronecker structure on\\
        \hline
         KRO-PRO-FAC& $Y_{i}$&OLS& mean \\
         KRO-PRO-FAC ($\alpha$)& rank$-\alpha$ $Y_{i}^{(\alpha)}$&OLS&mean \\
         rdu-rank-KRO ($\gamma$)& $Y_{i}$&  rank$-\gamma$ OLS $\tilde{\nu}^{(\gamma)}$&mean\\
         dual-KRO-MLE&$Y_{i}$& column \& row separate estimates& mean \& covariance\\
        \hline
    \end{tabular}
\end{table}
\begin{table}
 \caption{\label{tab:M1} Average relative estimation error ($\%$) under the identity covariance (Model 1)}
    \centering
\begin{tabular}{l|rrrrr}
\hline
\multirow{2}{*}{$\|\hat{\nu}-\nu\|_{F}/\|\nu\|_{F}$}&\multicolumn{5}{c}{sample size (n)}\\
\multirow{2}{*}{} & 200 & 400 & 1000 & 2000 & 3000 \\
\hline
KRO-PRO-FAC & 0.339 & 0.237 & 0.151 & 0.106 & 0.087 \\
KRO-PRO-FAC ($\alpha=2$) & 0.339 & 0.238 & 0.151 & 0.106 & 0.087 \\
rdu-rank-KRO ($\gamma=2$) & 63.998 & 63.998 & 63.997 & 63.997 & 63.997 \\
dual-KRO-MLE & 76.938 & 52.702 & 20.721 & 8.377 & 6.563 \\
\hline
\hline
\end{tabular}
\end{table}
\begin{table}
 \caption{\label{tab:M2}Average relative estimation error ($\%$) under the banded covariance with a bandwidth $b=5$ (Model 2)}
    \centering
\begin{tabular}{l|rrrrr}
\hline
\multirow{2}{*}{$\|\hat{\nu}-\nu\|_{F}/\|\nu\|_{F}$}&\multicolumn{5}{c}{sample size (n)}\\
\multirow{2}{*}{} & 200 & 400 & 1000 & 2000 & 3000 \\
\hline
KRO-PRO-FAC & 1.508 & 1.059 & 0.666 & 0.472 & 0.385 \\
KRO-PRO-FAC ($\alpha=2$) & 1.552 & 1.116 & 0.746 & 0.576 & 0.506 \\
rdu-rank-KRO ($\gamma=2$) & 63.963 & 63.997 & 63.999 & 63.998 & 63.998 \\
dual-KRO-MLE & 38.934 & 19.116 & 7.211 & 2.292 & 2.418 \\
\hline
\hline
\end{tabular}
\end{table}
\begin{table}
 \caption{\label{tab:M3}Average relative estimation error ($\%$) under the AR(1) setting with $\rho=0.9$ (Model 3)}
    \centering
\begin{tabular}{l|rrrrr}
\hline
\multirow{2}{*}{$\|\hat{\nu}-\nu\|_{F}/\|\nu\|_{F}$}&\multicolumn{5}{c}{sample size (n)}\\
\multirow{2}{*}{} & 200 & 400 & 1000 & 2000 & 3000 \\
\hline
KRO-PRO-FAC & 0.351 & 0.247 & 0.157 & 0.110 & 0.090 \\
KRO-PRO-FAC ($\alpha=2$) & 0.512 & 0.440 & 0.391 & 0.373 & 0.371 \\
rdu-rank-KRO ($\gamma=2$) & 63.998 & 63.998 & 63.997 & 63.997 & 63.997 \\
dual-KRO-MLE & 0.250 & 0.177 & 0.113 & 0.079 & 0.064 \\
\hline
\hline
\end{tabular}
\end{table}
\begin{table}
 \caption{\label{tab:M4}Average relative estimation error ($\%$) under the Student's t-distribution with $5$ degrees of freedom (Model 4)}
    \centering
\begin{tabular}{l|rrrrr}
\hline
\multirow{2}{*}{$\|\hat{\nu}-\nu\|_{F}/\|\nu\|_{F}$}&\multicolumn{5}{c}{sample size (n)}\\
\multirow{2}{*}{} & 200 & 400 & 1000 & 2000 & 3000 \\
\hline
KRO-PRO-FAC & 0.439 & 0.308 & 0.194 & 0.137 & 0.112 \\
KRO-PRO-FAC ($\alpha=2$) & 0.440 & 0.308 & 0.195 & 0.137 & 0.113 \\
rdu-rank-KRO ($\gamma=2$) & 63.998 & 63.998 & 63.997 & 63.997 & 63.997 \\
dual-KRO-MLE & 69.049 & 39.734 & 12.677 & 7.141 & 5.010 \\
\hline
\hline
\end{tabular}
\end{table}

For Model 1 we see from Table~\ref{tab:M1} that both the KRO-PRO-FAC
and KRO-PRO-FAC ($\alpha$) method have the smallest estimation
error. The dual-KRO-MLE estimate is substantially less accurate compared to that of
KRO-PRO-FAC and KRO-PRO-FAC ($\alpha$) especially when the sample size
is small, e.g., $n = 200$ or $n = 400$. This is in a sense expected as
the entries of $E_i$ are iid and thus 
there are few if any benefits in estimating and/or incorporating the
covariance structure of $\{Y_i\}$. 
Finally, the rdu-rank-KRO ($\gamma$) method has the highest estimation
error and this observation also extends to the results for Model $2$
through $4$ as presented in Table~\ref{tab:M2} through Table~\ref{tab:M4}
This is once again expected as, recalling the earlier discussions in
Example~\ref{eg: example1}, the Kronecker structure in the regression
coefficient $\nu$ is fundamentally different from assuming $\nu$ to be
low-rank. In other words imposing rank constraints on $\tilde{\nu}$
only leads to information loss due to model
misspecification. 

For Model~2 we see from Table~\ref{tab:M2} that the KRO-PRO-FAC algorithm
has the smallest estimation error with the KRO-PRO-FAC ($\alpha$)
variant being slightly worse. The estimate obtained from the dual-KRO-MLE algorithm is 
noticably worse compared to both the KRO-PRO-FAC and KRO-PRO-FAC
($\alpha$) and furthermore appeared to be sensitive to the sample size
$n$, i.e., its estimation error is much larger than its competitors
when $n = 200$ or $n = 400$. We note that for this Model $2$, $90\%$ of the non-zero correlations
in $\Sigma_{\vect{(E)}}$ have absolute value less than $0.5$, which
suggests either weak or mild dependence among rows and columns in $\vect{(Y_{i})}$. 

For Model~3 we see from Table~\ref{tab:M3} that the dual-KRO-MLE
algorithm yields the most accurate estimates with errors that are
slightly smaller than that of KRO-PRO-FAC and KRO-PRO-FAC
($\alpha$) methods. There is thus value in joint modeling of the mean $\nu$
and the covariance structure for the $\{Y_i\}$. Note, however, that
the KRO-PRO-FAC algorithm is much less computationally demanding
compared to dual-KRO-MLE. 

Finally, for Model $4$ we see from Table~\ref{tab:M4} that the
KRO-PRO-FAC algorithm outperforms all of its competitors. In
particular it is slightly better than KRO-PRO-FAC($\alpha$) and is
much better than dual-KRO-MLE. These results are similar to that in Table~\ref{tab:M1}
and one possible explanation for this similarity is that both models
induce the same covariance structure for $\{Y_i\}$. 

\subsection{Real data analysis}
\label{subsec:real_data}
We now apply the KRO-PRO-FAC algorithm to the electroencephacology
(EEG) dataset from the \href{https://archive.ics.uci.edu/ml/datasets/EEG+Database}{UC Irvine
Machine Learning Repository}. The data arises from a study of EEG
measurements related to alcoholoism in which there are $122$ subjects
from either the alcoholic group ($77$ subjects) or the control group
($45$ subjects). For each subject a series of voltage measurements is
made at $256$ different time points from $64$ different regions of the
scalp, i.e., the EEG response for the $i$th subject in the $j$th group
(with $j = 1$ and $j = 2$ denoting the alcoholic and control) can be viewed as
a matrix $Y_{ij}$ with $256$ rows and $64$ columns. A key research
question for this dataset is to identify which of the $64$ brain channel accounts for most of the
differences in voltages measurements between the two groups.

To answer the above inquiry we partition the data according to the subject
grouping and fit a bi-linear model of the form Eq~\eqref{eq: bi-linear form
  each d sum} to each group. As the EEG dataset contains no other
covariates, this lead to a model 
of the form
\begin{align}
\label{eq: eeg d sum}
    \vect(Y_{ij}) &= \big(\sum_{k=1}^{d^{(j)}}\beta_{2k}^{(j)} \otimes \beta_{1k}^{(j)}\big)   + \vect(E_{ij}),\quad i\in[n_{j}],\,j=1,2 
\end{align}
where $\beta_{2k}^{(j)}\in\Re^{64\times1}$ and
$\beta_{1k}^{(j)}\in\Re^{256\times1}$. In other words, the mean
response $\nu^{(j)} = \mathbb{E}[\vect{(Y_{ij})}]$ for the $j$th group is a sum
of $d^{(j)}$ Kronecker products and thus $\nu^{(1)} - \nu^{(2)}$ is
the effect of alcoholism (when compared to the control group) on the
voltage measurements. 
We emphasize that the number of Kronecker
factors $d^{(j)}$ are possibly different between the two groups.
We apply the KRO-PRO-FAC algorithm to these $\{Y_{ij}\}$ with
$d^{(1)} = 2$ and $d^{(2)} = 3$ chosen via the singular
value ratio criterion as described in Eq.~\eqref{eq: choose_d}. Let
$\hat{\nu}^{(1)} = \sum_{k=1}^{2} \hat{\beta}_{2k}^{(1)} \otimes
  \hat{\beta}_{1k}^{(1)}$ and 
$\hat{\nu}^{(2)} = \sum_{k=1}^{2} \hat{\beta}_{2k}^{(2)} \otimes
  \hat{\beta}_{1k}^{(2)}$ be the resulting estimates of $\nu^{(1)}$
  and $\nu^{(2)}$.

Given these $\hat{\nu}^{(1)}$ and $\hat{\nu}^{(2)}$ we then follow the
  same post-processing steps described in \cite{Envelope_Models} for
  multiple testing among the brain locations.
  Firsly, we isolate the alcoholism effects of each channel by
  averaging out the time effects $\hat{\nu}^{(1)} - \hat{\nu}^{(2)}$,
  ie.,  we take the {\em column}
  means of the $\vect^{-1}{(\hat{\nu}^{(1)}-\hat{\nu}^{(2)},256,64)}$ where $\vect^{-1}{(\cdot,256,64)}$ yields a matrix of
  dimensions $256 \times 64$. This yields in a vector $\hat{\theta}
  \in \Re^{64}$ which we then 
 conduct multiple t-test for the null hypothesis that $\mathbb{H}_0
 \colon \theta_i = 0$ and compute the resulting p-values. Finally we
 apply the Benjamini–Yekutieli procedure \citep{BY_p-values} to adjust
 these $p$-values. 

\begin{figure}
\centering
\includegraphics[width=0.85\linewidth]{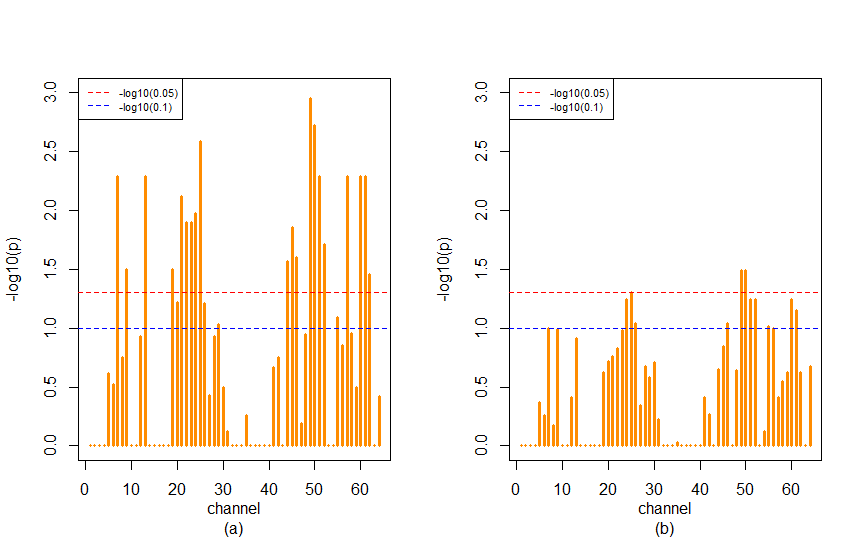}
\caption{\label{fig:eeg_pvalue} Benjamini–Yekutieli adjusted p-values
  (on a scale of $-\log_{10}$) for $64$ brain channels obtained from the (a) Kronecker products estimates in Eq.~\eqref{eq: eeg d sum} with $d=2$ and $d=3$ for the
alcoholic and control group (b) OLS estimates}
\end{figure}

 The left panel of Figure~\ref{fig:eeg_pvalue} reports these adjusted
 p-values (on a $\log_{10}$ scale). For
 comparisons we also repeat the same post-processing analysis but replaced the
 estimates $\hat{\nu}^{(1)}$ and $\hat{\nu}^{(2)}$ by the
 the OLS estimate $\bar{Y}^{(1)} = n_1^{-1} \sum_{i \in n_1} Y_{i1}$ and $\bar{Y}^{(2)}
 = n_2^{-1} \sum_{i \in n_2} Y_{i2}$ and present the adjusted
 $p$-values for these OLS estimates in the right panel of
 Figure~\ref{fig:eeg_pvalue}. Figure~\ref{fig:eeg_pvalue} indicates that, for a significant
 level of $0.05$, the KRO-PRO-FAC estimates lead to the 
 detectation of $20$ possibly relevant channels while the OLS
 estimates detect only $3$ possibly relevant channels. We note that
 \cite{tensor_regression,Envelope_Models} also analyzed the same data
 set and their estimates detect ${\color{red} 24}$ and ${\color{blue}
   26}$ possibly relevant channels, respectively. Our detections using 
 the KRO-PRO-FAC estimates are thus comparable with those
 from \cite{tensor_regression,Envelope_Models}; indeed they all
 detected the regions from $21$ to $25$, from $44$ to $52$ and from
 $57$ and $62$. The main benefit of using the KRO-PRO-FAC estimates is
 that they can be commputed efficiently and do not
 depend on knowing or estimating $\mathrm{Cov}[\vect{(Y_{ij})}]$.

\section{Conclusion}
\label{sec:conc}
In this paper we consider matrix regression $Y_i =
\sum_{k} \beta_{1k} X_i \beta_{2k}^{\top} + E_i$ where the responses
$Y_i$ are high-dimensional matrices and propose a computationally efficient procedure
for estimating $\{\beta_{1k}, \beta_{2k}\}$ based on the nearest
Kronecker products approximation to the OLS estimate $\hat{\nu}$ of
$\nu = \sum_{k} \beta_{2k} \otimes \beta_{1k}$. 
We now mention three potential directions for future research. The
empirical results in Section~\ref{subsec:simulation} show that the
KRO-PRO-FAC procedure has smallest estimation error when the noise
entries for
$E_i$ are independent and is slightly worse than the
dual-KRO-MLE procedure of \cite{Envelope_Models} when the noise of
$E_i$ are highly correlated. As the dual-KRO-MLE is somewhat
computationally demanding, it will be valuable to refine our
KRO-PRO-FAC procedure for handling highly dependent rows and columns
without compromising its computation efficiency. Secondly, the
performance of the low-rank variant KRO-PRO-FAC ($\alpha$) is 
also quite competitive but its theoretical property is currently
unaddressed. Finally, for many type of matrix data such as those
arising in image analysis, the ordering of the rows and columns for these matrices
are based on latent but important features. For example, pixels'
intensities in an image usually exhibit some continuity in both 
vertical and horizontal directions. How to meaningfully 
extract these latent features and incorporate them into the matrix
regression problem is an open and interesting research question. 
\bibliographystyle{rss}
\bibliography{example}
\appendix
\section{Proofs of Stated Results}\label{appendix}
\subsection{{Proof of Theorem~\ref{thm: sin_theta}} }
For convenience of notation, we take 
\begin{align}
    \underbrace{(\mathcal{X}^{\top}\mathcal{X})^{-1}\mathcal{X}^{\top}}_{q_{1}q_{2} \times n}&=\begin{bmatrix}
       \mathcal{C}_{1},\mathcal{C}_{2},\dots,\mathcal{C}_{q_{1}q_{2}}
    \end{bmatrix}^{\top},\: \mathcal{C}_{1},\mathcal{C}_{2},\dots,\mathcal{C}_{q_{1}q_{2}} \in \Re^{n}\\
    \underbrace{\mathcal{E}}_{n\times p_{1}p_{2}}&=\begin{bmatrix} 
   \vect(E_1)^{\top} \\
    \vdots \\
    \vect(E_n)^{\top}\\ 
    \end{bmatrix}=\begin{bmatrix}
       \mathcal{D}_{1},\mathcal{D}_{2},\hdots,\mathcal{D}_{p_{1}p_{2}}
    \end{bmatrix}=\begin{bmatrix}
       \mathcal{F}_{1},\mathcal{F}_{2},\hdots,\mathcal{F}_{p_{1}p_{2}}
    \end{bmatrix}\Sigma_{\vect{(E)}}^{1/2}
\end{align}
where $\: \mathcal{D}_{1},\mathcal{D}_{2}\cdots\mathcal{D}_{p_{1}p_{2}} \in \Re^{n}$ and $\: \mathcal{F}_{1},\mathcal{F}_{2}\cdots\mathcal{F}_{p_{1}p_{2}} \in \Re^{n}$ with $\mathcal{F}_{s}=(\xi_{1s},\dots,\xi_{ns})^{\top}$ being independent with independent entries. 

Set
$\Delta=\left[(\mathcal{X}^{\top}\mathcal{X})^{-1}\mathcal{X}^{\top}\mathcal{E})\right]^{\top}$
and recall that
$R(\nu)=\sum_{k=1}^{d}\vect{(\beta_{2k})}\vect{(\beta_{1k})}^{\top}=\sum_{k=1}^{d}\sigma_{k}\;\mathcal{U}_{k}\mathcal{V}_{k}^{\top}$. We
then have
\begin{align*}
    R(\hat{\nu})  &=\sum_{k=1}^{d}\sigma_{k}\:\mathcal{U}_{k}\mathcal{V}_{k}^{\top}+R(\Delta)
\end{align*}
By Weyl's inequality (Problem~III.6.13 in \citet{bhatia2013matrix})
and Wedin $\sin$-$\Theta$ \citep{wedin_perturbation} theorem we have
\begin{align}
    |\hat{\sigma}_{k}-\sigma_{k}|&\leq \bigl\|R(\Delta)\bigr\|\\
    \lVert \sin  \Theta\:(\mathcal{U}_{1},\hat{\mathcal{U}}_{1})\rVert_{2},\, \lVert \sin \Theta\:(\mathcal{V}_{1},\hat{\mathcal{V}}_{1})\rVert_{2}&\leq \frac{\min\{\lVert\hat{\mathcal{U}}_{1}^{\top}R(\Delta)\rVert_{2},\lVert R(\Delta) \hat{\mathcal{V}}_{1}\rVert_{2}\}}{\hat{\sigma}_{d}}\leq\frac{\lVert R(\Delta)\rVert_{2}}{\hat{\sigma}_{d}}
\end{align}
It thus suffices to bound the spectral norm of $R(\Delta)$. First note
that $R(\Delta)$ can be written as a block matrix of the form
$$R(\Delta) = \begin{bmatrix} \tilde{\Delta}_{11} &
                                                    \tilde{\Delta}_{12}
                & \dots & \tilde{\Delta}_{1 q_1} \\
                \tilde{\Delta}_{21} & \tilde{\Delta}_{22} & \dots &
                                                            \tilde{\Delta}_{2q_1} \\
                \vdots & \vdots & \ddots & \vdots \\
                \tilde{\Delta}_{q_2 1} & \tilde{\Delta}_{q_2 2} &
                                                                  \dots &\tilde{\Delta}_{q_2 q_1}
                \\
              \end{bmatrix}, \quad \tilde{\Delta}_{k \ell} =
    \begin{bmatrix}
      \mathcal{D}_{1}^{\top}\mathcal{C}_{(k-1)q_1 + \ell
      } &\dots
      &\mathcal{D}_{p_{1}}^{\top}\mathcal{C}_{(k-1)q_1 + \ell} \\
    \mathcal{D}_{p_{1}+1}^{\top}\mathcal{C}_{(k-1)q_1 +
      \ell}&\dots
      &\mathcal{D}_{2p_{1}}^{\top}\mathcal{C}_{(k-1)q_1 + \ell }&  \\
    \vdots&\ddots&\vdots\\
    \mathcal{D}_{(p_{2}-1)p_{1}+1}^{\top}\mathcal{C}_{(k-1)q_2 +
      \ell}&\dots
      &\mathcal{D}_{p_{1}p_{2}}^{\top}\mathcal{C}_{(k-1)q_1 + \ell} \end{bmatrix} $$
The matrix $\tilde{\Delta}_{k \ell}$ can be further expressed as
\begin{align}
\tilde{\Delta}_{k \ell} = \begin{bmatrix}
      \bm{\Sigma}_{1}^{1/2} \zeta_{1} &  \bm{\Sigma}_{2}^{1/2}
                                        \zeta_{2} & \dots &
      \bm{\Sigma}_{p_{2}}^{1/2} \zeta_{p_{1}} \end{bmatrix},
 \quad \zeta_{s}=[\mathcal{F}_{s},\mathcal{F}_{p_{1}+s},
           \mathcal{F}_{2p_{1}+s},\dots,\mathcal{F}_{(p_2 - 1)p_{1}+s}]^{\top} \mathcal{C}_{(k-1)q_1 + \ell}
\end{align}
Note that for ease of exposition we had suppressed the dependency on
$k$ and $\ell$ in the notation for $\zeta_{s}$. This should cause
minimal confusion as we can fix some arbitrary $k$ and $\ell$ before
proceeding with the subsequent derivations.

We now derive a concentration inequality for $\|\tilde{\Delta}_{k
  \ell} \|$ using a standard $\epsilon$-net argument. 

\textbf{Step 1: $\epsilon$ net} Let $\epsilon=1/4$ and choose an
$\epsilon$ net $\mathcal{M}$ of the sphere $\mathcal{S}^{p_{2}-1}$ and
an $\epsilon$ net $\mathcal{R}$ of the sphere
$\mathcal{S}^{p_{1}-1}$. We have
\begin{align*}
    |\mathcal{M}|\leq 9^{p_{2}},\quad |\mathcal{R}|\leq 9^{p_{1}}
\end{align*}
The spectral norm of $\tilde{\Delta}_{k \ell}$ can then be bounded as
\begin{align*}
    \lVert \tilde{\Delta}_{k \ell} \rVert_{2}\leq2\max_{x\in \mathcal{M},y\in\mathcal{R}}\langle
 \tilde{\Delta}_{k \ell} x,y\rangle
\end{align*}

\textbf{Step 2: Concentration} Fix $x \in \mathcal{M}$ and $y \in
\mathcal{R}$. We then have
\begin{align*}
    \langle
 \tilde{\Delta}_{k \ell} x,y\rangle=&\sum_{i=1}^{p_{2}} x_{i}\zeta_{i}^{\top}\bm{\Sigma}_{i}^{1/2}y
    = \sum_{i=1}^{p_{2}} x_{i}\big[\sum_{s=1}^{p_{1}}\sum_{k=1}^{p_{1}}\bm{\Sigma}_{i}^{1/2}(k,s)\zeta_{ik}y_{s}\big]
\end{align*}
Using properties of the Orlicz $\Psi_{2}$-norm (see e.g.,
Proposition~2.6.1 in \cite{HDProbability}) we have
\begin{align*}
    \lVert\langle
 \tilde{\Delta}_{k \ell}x,y\rangle\rVert_{\psi_{2}}^{2}\leq&\mathcal{C}\sum_{i=1}^{p_{2}} x_{i}^{2}\big\lVert\sum_{s=1}^{p_{1}}\sum_{k=1}^{p_{1}}\bm{\Sigma}_{i}^{1/2}(k,s)\zeta_{ik}y_{s}\big\rVert_{\psi_{2}}^{2}\\
 \leq&\mathcal{C}\sum_{i=1}^{p_{2}} x_{i}^{2}\sum_{s=1}^{p_{1}}y_{s}^{2}\sum_{k=1}^{p_{1}}\big\lVert\bm{\Sigma}_{i}^{1/2}(k,s)\zeta_{ik}\big\rVert_{\psi_{2}}^{2}\\
 \leq&\mathcal{C}\max_{i,k}\lVert\zeta_{ik}\rVert_{\psi_{2}}^{2}\sum_{i=1}^{p_{2}} x_{i}^{2}\sum_{s=1}^{p_{1}}y_{s}^{2}\sum_{k=1}^{p_{1}}\big|\bm{\Sigma}_{i}^{1/2}(k,s)\big|^{2}\\
 \leq&\mathcal{C}\max_{i,k}\lVert\zeta_{ik}\rVert_{\psi_{2}}^{2}\max_{i\in [p_{2}]}\max_{s \in[p_{1}]}\bm{\Sigma}_{i}(s,s)\\
 \leq&\mathcal{C}\lVert(\mathcal{X}^{\top}\mathcal{X})^{-1}\mathcal{X}\rVert_{2}\max_{i\in [m]}\max_{s \in[r]}\bm{\Sigma}_{i}(s,s)
\end{align*}
where the second to last inequality is because
\begin{align*}
    \sum_{k=1}^{p_{1}}\big|\bm{\Sigma}_{i}^{1/2}(k,s)\big|^{2}
  = \sum_{k=1}^{p_{1}}\bm{\Sigma}_{i}^{1/2}(s,k)\bm{\Sigma}_{i}^{1/2}(k,s)
    = \bm{\Sigma}_{i}(s,s).
\end{align*} 
Let $K =
\lVert(\mathcal{X}^{\top}\mathcal{X})^{-1}\mathcal{X}\rVert \times \max_{i\in
  [m]}\max_{s \in[r]}\bm{\Sigma}_{i}(s,s)$. We therefore have, for all
$u \geq 0$, that
\begin{align}
  P(\langle \tilde{\Delta}_{k \ell} x,y\rangle \geq u)\leq 2 \exp{(-cu^{2}/K^{2})}.
\end{align}

\textbf{Step 3: Union bound} By union over the $\mathcal{M}$ and $\mathcal{R}$, then with probability $1-2\exp{(-u^2)}$, we have for any $u>0$
\begin{align}\label{eq: Aw tail bound}
    \lVert \tilde{\Delta}_{k \ell} \rVert \leq \mathcal{C} \lVert(\mathcal{X}^{\top}\mathcal{X})^{-1}\mathcal{X}\rVert_{2}\max_{i\in [p_{2}]}\max_{s \in[p_{1}]}\bm{\Sigma}_{i}(s,s)(\sqrt{p_{2}}+\sqrt{p_{1}}+u).
\end{align}
This upper bound is independent of $\{k,\ell\}$ and since
$\|R(\Delta)\| \leq \sum_{k=1}^{q_{1}} \sum_{\ell=1}^{q_{2}}\lVert
\tilde{\Delta}_{k \ell} \rVert$, we obtain the desired results in
Theorem~\ref{thm: sin_theta}.
\subsection{{Proof of Corollary~\ref{prop: Kronecker}} }
We will continue to use the same notations as that in the proof of Theorem~\ref{thm: sin_theta}. 
Let $\mathcal{P}_0 = \mathcal{U} \mathcal{U}^{\top}$ and
$\mathcal{P}_1 = \mathcal{V} \mathcal{V}^{\top}$. Similarly, 
let $\hat{\mathcal{P}}_{0} = \hat{\mathcal{U}} \hat{\mathcal{U}}^{\top}$
and $\hat{\mathcal{P}}_{1} = \hat{\mathcal{V}}
\hat{\mathcal{V}}^{\top}$. Note that these matrices are all of rank at
most $d$. As $R(\nu) = \mathcal{P}_0 R(\nu) \mathcal{P}_1$, we have
$$\hat{\mathcal{P}}_0 R(\tilde{\nu}) \hat{\mathcal{P}}_1 - R(\nu) =
(\hat{\mathcal{P}}_0 - \mathcal{P}_0) R(\tilde{\nu}) \hat{\mathcal{P}}_1 + \mathcal{P}_0
R(\tilde{\nu}) (\hat{\mathcal{P}}_1 - \mathcal{P}_1) + \mathcal{P}_0
R(\Delta) \mathcal{P}_1.$$
We therefore have
$$\|\hat{\mathcal{P}}_0 R(\tilde{\nu}) \hat{\mathcal{P}}_1 -
R(\nu)\|_{F} \leq \sqrt{d} \|\hat{\mathcal{P}}_0 - \mathcal{P}_0\|_{2} \times
\|R(\tilde{\nu})\|_{2} + \sqrt{d} \|\hat{\mathcal{P}}_1 - \mathcal{P}_1\|_{2}
\times \|R(\tilde{\nu})\|_{2} + \sqrt{d} \|R(\Delta)\|_{2}.$$

Now $\|\hat{\mathcal{P}}_0 - \mathcal{P}_0\|_{2} \leq 2\|\sin
\Theta(\hat{\mathcal{U}}, \mathcal{U})\|_{2}$ and similarly for
$\|\hat{\mathcal{P}}_1 - \mathcal{P}_1\|_{2}$. Then from the conditions in
Assumption~\ref{ass: asymptotics}, we have
$$\|\hat{\mathcal{P}}_0 R(\tilde{\nu}) \hat{\mathcal{P}}_1 -
R(\nu)\|_{F} \leq 2 \sqrt{d} \|\sin \Theta(\hat{\mathcal{U}},
\mathcal{U})\|_{2} \times  (\|R(\nu)\|_{2} + \|R(\Delta)\|_{2}) + \|R(\Delta)\|_{2} =
\mathcal{O}(n^{-1/2} p^{1/2})$$
Finally, as $\hat{\nu}$ and $\nu$ are the {\em inverse} rearrangement of
$\hat{\mathcal{P}}_0 R(\tilde{\nu}) \hat{\mathcal{P}}_1$ and $R(\nu)$,
respectively, we have
$$\frac{\|\hat{\nu} - \nu\|_{F}}{\|\nu\|_{F}} =  \frac{\|\hat{\mathcal{P}}_0 R(\tilde{\nu}) \hat{\mathcal{P}}_1 -
R(\nu)\|_{F}}{\|\nu\|_{F}} =
\mathcal{O}((np)^{-1/2})$$
as desired.

\end{document}